\documentclass[letterpaper]{article}

\usepackage{aaai19}

\usepackage{times}
\usepackage{indentfirst}
\usepackage{helvet}
\usepackage{courier}
\usepackage{graphicx}
\usepackage{hyperref}
\usepackage{amsmath}
\usepackage{amsfonts}
\usepackage{here}
\usepackage{multirow}
\usepackage{colortbl, array}
\usepackage{bm} 
\usepackage{url}
\usepackage[bottom]{footmisc}

\DeclareMathOperator*{\argmax}{argmax}
\newcommand{\textqcr}[1]{{\fontfamily{qcr}\selectfont #1}}

\begin{document}

\title{An Ensemble Dialogue System for Facts-Based Sentence Generation}

\author{Ryota Tanaka, Akihide Ozeki, Shugo Kato, Akinobu Lee\\
Graduate School of Engineering, Nagoya Institute of Technology, Japan\\
\{rtanaka, ozeki, shugo693, ri \}@slp.nitech.ac.jp
}

\maketitle
\begin{abstract}
\begin{quote}
This study aims to generate responses based on real-world facts by conditioning context and external facts extracted from information websites. Our system is an ensemble system that combines three modules: generated-based module, retrieval-based module, and reranking module. Therefore, this system can return diverse and meaningful responses from various perspectives. The experiments and evaluations are conducted with the sentence generation task in Dialog System Technology Challenges 7 (DSTC7-Task2). As a result, the proposed system performed significantly better than sole modules, and worked fine at the DSTC7-Task2, specifically on the objective evaluation.
\end{quote}
\end{abstract}

\section{Introduction}
The popularization of Social Networking Services (SNS) offers the advantage of reducing the burden of building large-scale open datasets. Therefore, recent works pertaining to dialogue systems have focused on end-to-end dialogue system using neural networks \cite{NCM,HRED,DBLP:conf/aaai/SerbanSLCPCB17}. The end-to-end approach has a potential to generate tailored and coherent responses for user-input. However, there are still some problems with suffering from ``safe response'' phenomenon available to any utterance, such as the ``\textit{I'm sorry}''and ``\textit{I think so},'' and generating words that have meanings different from real-world facts. This is because neural networks generally infer responses using only the collection of conversational transcriptions.

To tackle these problems, researchers have taken various approaches. \cite{DBLP:conf/aaai/GhazvininejadBC18} proposed a knowledge-grounded dialogue system, conditioned on the context and facts extracted from online resources such as SNS posts utilizing location information. This easily and quickly enables to handle topics not appeared in training data and to adapt to a new domain. In the other approach, dialogue systems combining multiple dialogue models allow responses to be more diverse than those with a single model so that they can treat user-inputs from various viewpoints \cite{DBLP:journals/corr/abs-1709-02349,DBLP:conf/ijcai/SongLNZZY18}. We believe that combining these approaches is crucial to generate meaningful responses. 

In this study, we propose an ensemble dialogue system conditioned on a previous context and external facts. This system consists of three modules including generation, retrieval and reranking. First, two modules generate and retrieve responses by feeding context and facts extracted from information websites such as Wikipedia. In generating candidates, we use the method extending Diverse Beam Search (DBS) \cite{DBS} by enhancing the probability of words in facts data to treat low-frequency words such as proper nouns in external data adequately. Second, the reranking module sorts these candidates according to several features considering appropriateness and informativeness, and it finally returns the final response which is the highest-ranked candidate. Our main contributions of this paper has two-fold : (1) we propose a model for combining multiple hypotheses and injecting external facts, (2) we develop a method to decode diverse and informative words. 

We evaluate the performance with the DSTC7-Task2 \cite{DSTC7}, which is devoted to building dialogue systems generating responses based on real-world facts. In this paper, we report our experimental results.

\section{Problem Definition}
The system outputs a response using a context $S = \{ U_1,...,U_M\}$ in $M$ recent turns and $N$ facts $F = \{f_1,...,f_N\}$ relevant to the context, where $F$ is a sentence, containing HTML tag, extracted information websites. Each utterance  $U_m = \{x_{m,1},...,x_{m,n}\}$ is composed of $n$ words.

Here, we categorize $F$ as subject facts $F^{subj}=\{f^{subj}_1,...,f^{subj}_K\}$ and description facts $F^{desc}=\{f^{desc}_1,..., f^{desc}_L\}$ using the HTML tag rule. $F^{subj}$ is a sequence enclosed by \verb|<h>| tag or \verb|<title>| tag, and $F^{desc}$ is a sequence enclosed by \verb|<p>| tag or not enclosed by any tags.

\section{Ensemble Dialogue System for Facts-Based Sentence Generation}
\begin{figure}[t]
    \centering
    \includegraphics[scale=0.3]{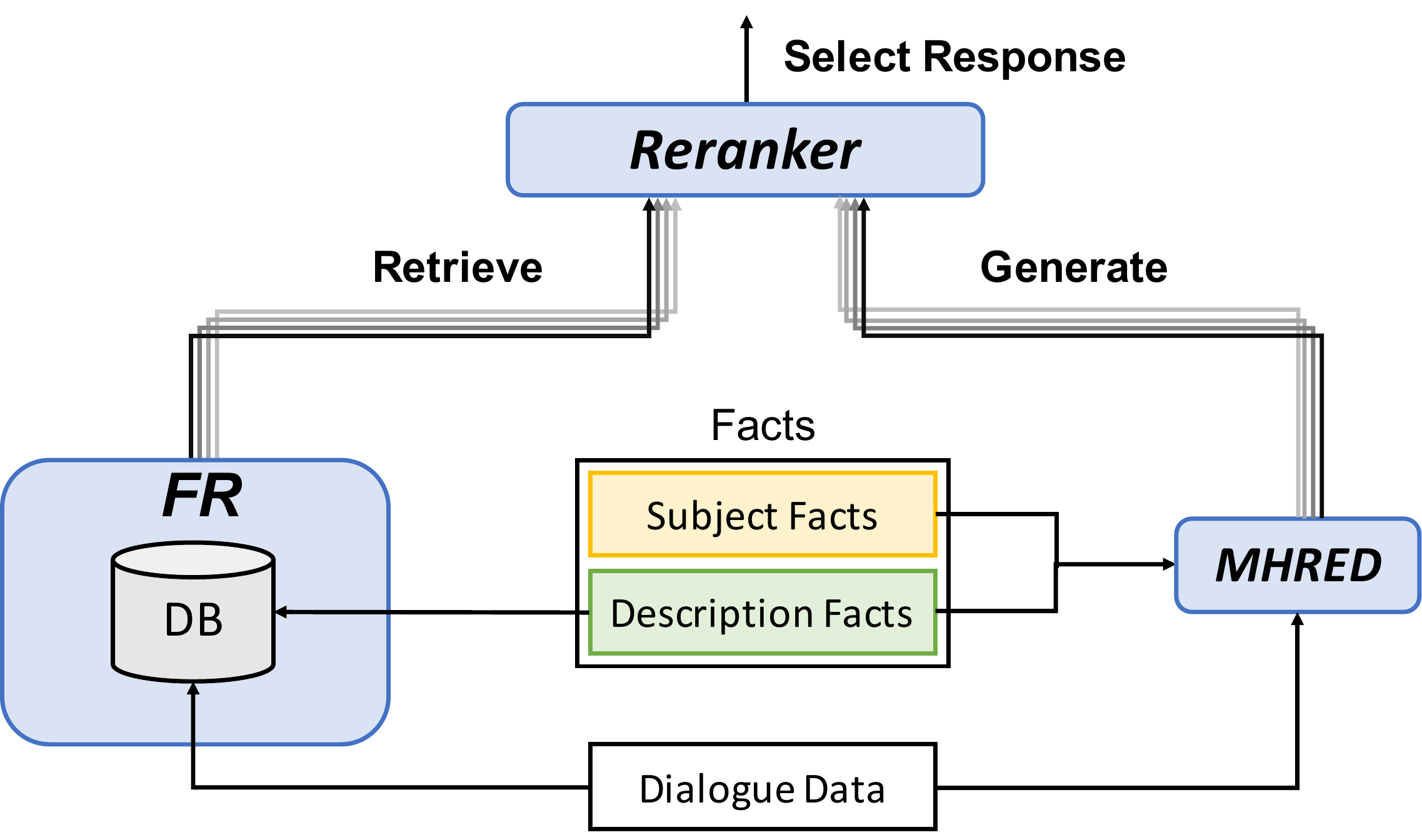}
    \caption{An overview of the proposed model}
    \label{fig:model_en}
\end{figure}

We propose an ensemble dialogue system using external facts and context. As shown in Figure \ref{fig:model_en}, it consists of the Memory-augmented Hierarchical Encoder-Decoder (MHRED), the sentence selection module with facts retrieval (FR), and the Reranker. This system has two processes: the generate-retrieval process and the reranking process. In the generate-retrieval process, the MHRED generates responses using context and external facts, and the FR retrieves the responses from a database containing important words extracted from the facts. In the reranking process, we use a binary classifier with various dialogue features to select the final response by feeding all the candidates from the MHRED and FR. In this section, each module of the proposed system is introduced in detail.

\subsection{Memory-augmented Hierarchical Recurrent Encoder-Decoder}

\begin{figure}[t]
    \centering
    \includegraphics[scale=0.3]{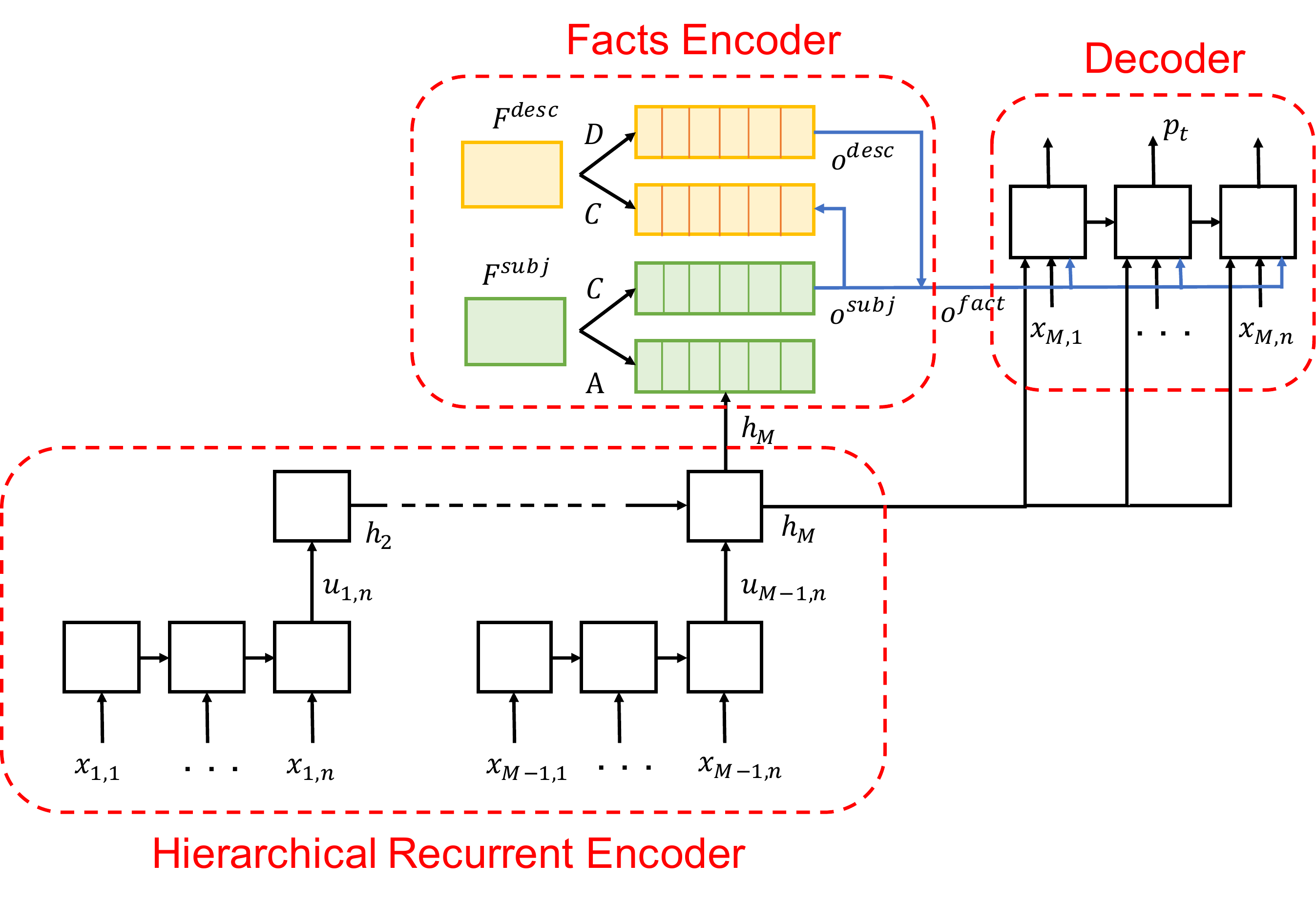}
    \caption{An overview of the MHRED}
    \label{fig:mhred}
\end{figure}

To inject facts into responses, a novel encoder-decoder model incorporating end-to-end memory networks (MemN2N) \cite{MemN2N} architecture into hierarchical recurrent encoder-decoder (HRED) \cite{HRED} is proposed.  We call this model as Memory-augmented HRED (MHRED). The overview of MHRED is shown in Figure \ref{fig:mhred}.

\subsubsection{Hierarchical Recurrent Encoder}
To encode the context, a Hierarchical Recurrent Encoder (HRE) is applied. Previous work has shown that hierarchical Recurrent Neural Networks (RNNs) have a higher ability to express the dialogue context than non-hierarchical RNNs \cite{P17-2036}. The HRE consists of two level encoders, one at the utterance level and the other at the context level, computed by the Gated Recurrent Unit (GRU) \cite{GRU}. An utterance encoder converts each utterance to an utterance vector. The utterance vector is the hidden state obtained after encoding the last word in each utterance. Let $ w_{m,t}$ denote the word embedding of the $t$-th word in the $m$-th utterance. Then, utterance vector $u_{m,t}$ is computed as follows: 
\begin{eqnarray}
u_{m,t} = \text{GRU} (u_{m,t-1}, w_{m,t}) 
\end{eqnarray}

After processing each utterance, a context encoder outputs context vector $h_m$, which is a summary of the past utterances, as follows:
\begin{eqnarray}
h_m &=& \text{GRU}(h_{m-1}, u_{m-1,n})
\end{eqnarray}

\subsubsection{Facts Encoder}
A facts encoder is introduced to select facts that need to be injected in responses and map the facts to the continuous representation utilizing the concept of MemN2N architecture. $F^{subj}$ contains many sentences, written headlines, and titles concerning facts, whereas $F^{desc}$ mostly contains sentences explaining the headline and the title. To access the $F^{desc}$ using $F^{subj}$, it is efficient to extract the detailed facts about the headlines and titles since they tend to contain vital information as a fact. 
Therefore, we extend the facts encoder proposed by \cite{DBLP:conf/aaai/GhazvininejadBC18} and to store $F^{subj}$ in the first memory (first hop), and $F^{desc}$ in the last memory (second hop).

First, $F^{subj}$ and $F^{desc}$ are converted into memory vector $r^{subj} = \{r^{subj}_1,...,r^{subj}_K\}, r^{desc} = \{ r^{desc}_1,...,r^{desc}_L \}$ by sum of word embeddings for each sentence. Then, context vector $h_M$, which is the last hidden state of HRE, is fed into the facts encoder in the first memory, and subject fact $o^{subj}$ is obtained, as shown below:
\begin{eqnarray}
m^{subj}_i &=& A r^{subj}_i \\
c^{subj}_i &=& C r^{subj}_i \\
p^{subj}_i &=& {\rm softmax} (h_M ^\mathrm{T} m^{subj}_i ) \\
o^{subj} &=& \sum^K_i p^{subj}_i c^{subj}_i 
\end{eqnarray}
where $A, C\in \mathbb{R}^{d \times |V|}$ ($|V|$ denotes the vocabulary size) are trainable parameters.
Moreover, $h_M$ and $o^{subj}$ are passed to the second memory, and we obtain the description fact $o^{subj}$ as follows:  
\begin{eqnarray}
m^{desc}_i &=& C r^{desc}_i \\
c^{desc}_i &=& D r^{desc}_i \\
p^{desc}_i &=& {\rm softmax}\{ (h_M + o^{subj})^\mathrm{T} m^{desc}_i \} \\
o^{desc} &=& \sum^L_i p^{desc}_i c^{desc}_i 
\end{eqnarray}
where $C, D\in \mathbb{R}^{d \times |V|}$ are trainable parameters. Note that $C$ denotes the shared weights between memories. Finally, vector concatenation across the rows on $o^{subj},o^{desc}$ is performed and facts vector $o^{fact}=[o^{subj};o^{desc}]$ is obtained.

\subsubsection{Decoder}
A decoder reads context vector $h_M$ and facts vector $o^{fact}$ and predicts the next utterance. Let the initial hidden state be $s_0 =h_{M}$. Then, the hidden state of decoder $s_t$ is computed by GRU as follows:
\begin{eqnarray}
s_t = \text{GRU} (s_{t-1}, w_{M,t}) 
\end{eqnarray}
In generating conversational responses such as ``\textit{I think}'' and ``\textit{I know}'' on the decoder, it is not always necessary to use facts relevant to the context at all time steps. Hence, the decoder should change the preference to whether facts or other information needs to be used. We use Maxout Networks (MN) \cite{DBLP:conf/icml/GoodfellowWMCB13} to generate the response injecting facts. MN obtains the vector with the maximum value $e$, where $e$ can be computed with linear transformation of input features. Its vector represents the most important features from among all features, and then enables the decoder to switch depending on whether only facts are required. The probability of generating the word $p_t$ is calculated as follows:
\begin{eqnarray}
z_t &=& W_z w_{M,t} + U_z h_{M} + V_z s_{t} + H_z o^{fact} \\
e_t &=& [\max \{ z_{t,2j-1} , z_{t,2j} \} ]^{\mathrm{T}} (j =1,...,d) \\
p_t &=& \text{softmax}(W_e e_t)
\end{eqnarray}
where $W_z, U_z, V_z, H_z\in \mathbb{R}^{2d \times d}$, and $W_e \in \mathbb{R}^{|V| \times d} $ are trainable parameters.

\subsubsection{Diverse Sentence Generation with Facts}
Most neural dialogue systems apply Beam Search (BS) to generate the optimal response \cite{NCM,HRED,DBLP:conf/aaai/SerbanSLCPCB17}. However, BS does not guarantee diversity for the final response because word sequences within the beam width closely resemble each other. In addition, words such as proper nouns, which often appear in facts data, tend to be less selective than general words appearing in dialogue data.

Previous work extended BS to focus on alleviating the diversity problem. \cite{DBS} proposed Diverse Beam Search (DBS), generating diverse word sequence alternatives to BS. Given a beam width $B$, groups $G$, and beam width in group $B' = B/G$, beam sets at time step $t$ are divided into $G$ subsets. The DBS selects the word $Y^{g}_t$ in order of $g=1,...,G$ for these subsets as follows:
\begin{eqnarray}
Y^{g}_t = \argmax \limits_{y^{g}_{1,[t]},...,y^{g}_{B',[t]}}
\sum \limits_{b \in [B']} \Theta_t (y^{g}_{b,[t]}) + \lambda \Delta_{div}
\end{eqnarray}
where $\lambda$ is the hyper-parameter, $\Theta$ is the log probability, and $\Delta_{div}$ is the penalty which is the hamming distance between the words selected in the other groups and $y_{b,[t]}^g$. Note that the DBS sets the penalty $\Delta_{div} = 0$ at $g=1$.

Furthermore, we extend the DBS to add a penalty with facts. In order to enhance the probability of generating the word sequence to contain words in facts data, we introduce a penalty $\Delta_{fact}$, using the similarity between facts and the sequence of candidate words. Let $\gamma$ be the hyper-parameter. The penalty term $\gamma \Delta_{fact}$ is added to the equation (15) when the word $Y^{g}_t$ is selected. Here, $\Delta_{fact}$ is calculated as follows:
\begin{eqnarray}
\Delta_{fact} = \frac{1}{K + L} \sum_{n=1}^{K+L} \text{Sim} (\sum_{i=1}^t w_i^g, \sum_{j=1}^{|f_n|} w^{f_n}_j)
\end{eqnarray}
where $w^g$, $w^f$ is the $Y^{g}$, $f$ of word embeddings computed by Word2Vec \cite{Word2Vec} respectively, and Sim($\cdot , \cdot$) denotes cosine similarity.

\begin{table*}[t]
    \centering
        \begin{tabular}{|c|c|c|} \hline
            \textbf{Category} & \textbf{Features} & \multicolumn{1}{c|}{\textbf{About Features}} \\ \hline \hline
            \multirow{4}{*}{Candidate} & Length & Number of characters, and words\\ \cline{2-3}
            & Fluency & $N$-gram $(N=2,3)$ language model\\ \cline{2-3}
            & POS & Number of nouns, verbs, adjectives, and adverbs \\ \cline{2-3}
            & Fact & Frequency of words appeared in $F^{subj}$ and $F^{desc}$ / number of words \\ \hline
            \multirow{9}{*}{Pair} & Word sim & Cosine similarity between one-hot vectors of words\\ \cline{2-3}
            & $N$-gram sim & Cosine similarity between $N$-gram $(N=2,3)$\\ \cline{2-3}
            & Length sim & Similarity \footnotemark[2] of number of characters, and words \\ \cline{2-3}
            & Embedding sim & Cosine similarity between vectors computed as the averaged Word2Vec 
            \\ \cline{2-3} 
            & Sentimental sim & Similarity \footnotemark[3] of semantic orientations \cite{emotional}  \\ \cline{2-3}
            & POS sim & Cosine similarity between BoW (nouns, verbs, adjectives, and adverbs) \\ \cline{2-3}
            & Proper Noun sim & Cosine similarity between BoW of proper noun types, extracted by NLTK \footnotemark[4]\\ \cline{2-3}
            & \multirow{2}{*}{Keyword sim} & Cosine similarity between the averaged Word2Vec of keywords \\ 
            & & extracted by RAKE algorithm \cite{RAKE}\\ \hline
            \multirow{2}{*}{Context} & \multirow{2}{*}{Topic sim} & Cosine similarity between topic vectors by feeding  \\
            & & a context and candidate to LDA model \cite{LDA}\\ \hline
        \end{tabular}
    \caption{Features used to select a response on the Reranker.}
    \label{features}
\end{table*}

\subsection{Sentence Selection with Facts Retrieval}
In general, the raw human-human conversation is highly fluent and rich in variety, and often contains a considerable amount of information about a specific topic in itself. 
Thus, in this study a method that combines utterance selection based on facts is also proposed.
Hence, Facts Retrieval (FR) is employed to output responses, including facts in responses and context. Let $S$ be the context and $R$ be the response. The database is constructed in the form of $<[S;R], R>$, where $[S;R]$ is a query, which is word sequence concatenation on $S$ and $R$, and $R$ is a system output. Note that the database is used from the training dialogue dataset. 

For sentence selection, we extract important words $Q$ from facts and feed them into the database. Here, $Q$ denotes overlapping words in $F^{subj}$, which contains titles and headlines. In order to eliminate noises and improve the quality of retrieval, $Q$ is restricted to word sequence that includes at least one noun, verb, adjective, and adverb. FR outputs $R$ satisfying the relation $Q \in [S;R]$. If multiple sentences satisfy the relation, FR reranks sentences using the score produced by BM25F \cite{BM25F} and outputs up to 10 sentences. Note that FR will not output sentences if the relation is unsatisfied.

\subsection{Reranker}
The outputs of the MHRED and FR modules may contain meaningless and non-fluent responses. Hence, these responses should be eliminated; the responses should be both appropriate and informative. The Reranker sorts candidates by feeding all of the results of the MHRED and FR, and the highest ranked candidate is returned to user as the final response. It classifies whether a candidate is ``positive'' or ``negative'' as a response, where the probability of being ``positive'' is computed as a confidence score from binary classification with XGBoost \cite{xgboost}. The features of the Reranker consist of three categories, ``Candidate'' (responses returned by both the FR and MHRED),  ``Pair'' (a pair of a previous utterance and ``Candidate''), and ``Context'' (a pair of a context and ``Candidate''), as shown in Table \ref{features}. These categories enable evaluation of the quality of the responses.

Pairs of a context and a response from the dialogue dataset was used to build the training dataset. Contexts and responses with a high ``response score'' (over 100) on Reddit \footnotemark[1] was chosen as positive examples. Then, negative examples are generated on those contexts according to one of the following rules:
\begin{itemize}
    \item A randomly selected response with a low ``response score'' (1 or less) from a dialogue on another topic.
    \item A response that swap words and eliminates some words randomly from a positive example.
    \item A response that matches both of above-mentioned descriptions.
\end{itemize}

As a result, the dataset contains 44,449 ``positive'' and ``negative'' examples respectively.

\footnotetext[1]{\url{https://www.reddit.com/}}
\footnotetext[2]{Let $|U|$ and $|S|$ be the length of the previous utterance and candidate sentence normalized 0 to 1 respectively. Then, the similarity is calculated as $1.0 - | \;|U| - |S|\; |$.}
\footnotetext[3]{Let $U_{sent}$ and $S_{sent}$ be the average of the semantic orientations of the previous utterance and candidate. Then, the similarity is calculated as $1.0 - | (U_{sent} - S_{sent}) / 2 |$.}
\footnotetext[4]{
\url{https://www.nltk.org/}}

\section{Experiments}
\subsection{Datasets}
The experiment was performed according to the regulations of DSTC7-Task2. We crawled the dialogue dataset from 178 subreddits (subsidiary threads or categories on Reddit). Markdown and special symbols were eliminated from the crawled dialogue dataset, and for the same context, the context-response pair of the highest ``response score'' was selected. We crawled the facts dataset from 226 information sharing websites, such as Wikipedia. The facts were categorized into $F^{subj}$ and $F^{desc}$ as mentioned above, up to the top 10 sentences with the highest cosine similarity for each context. For calculating similarity, the average of the Word2Vec output with 256 dimension was used. Note that the Word2Vec model was trained only on the official training datasets according to DSTC-Task2 regulations. The pre-processing described above leads to the formation of the dialogue and facts datasets, as shown in Table \ref{dataset}.

\begin{table}[t]
    \centering
        \begin{tabular}{cccc} \hline 
         \textbf{Dialogue Dataset} & \textbf{train} & \textbf{dev} & \textbf{test} \\ \hline 
         \# Dialogues & 832908 & 40932 & 13440 \\
         Avg. Turns & 4.72 & 4.80 &  4.02 \\
         Avg. Tokens/Utterance & 23.32 & 23.64 & 34.84 \\ \hline  \hline
         \textbf{Facts Dataset} &  \\ \hline
         Avg. Tokens/Sentence (s) & 3.86 & 3.61 & 3.30 \\
         Avg. Tokens/Sentence (d) &17.11 &16.67 & 15.63 \\
         \# Topics (s)  &27735 &1152 & 3047 \\
         \# Topics (d) &27645 &1121 & 3063 \\ \hline
        \end{tabular}
        \caption{Statics of pre-processed dataset.``s'' and ``d'' denotes subject facts and description facts , respectively.}
    \label{dataset}
\end{table}

\subsection{Evaluation Metrics}
Automatic evaluation and human evaluation for responses were conducted in DSTC7-task2 organizers. For automatic evaluation, two types of metrics are used; one is word-overlap metrics, including BLEU \cite{BLEU}, NIST \cite{NIST} and METEOR \cite{METEOR}, and the other is the diversity metric using div \cite{DBLP:journals/corr/LiGBGD15}. In human evaluation, human evaluates responses rated with score 1 (Strong Disagree) to 5 (Strong Agree) for Appropriateness and Informativeness using crowdsourcing.

\subsection{Models for Comparison}
Several models are evaluated to show the effectiveness of the proposed model:
\begin{itemize}
    \item \textqcr{S2S}: Sequence-to-sequence (seq2seq) model \cite{NCM}.
    \item \textqcr{HRED}: HRED model \cite{HRED}. 
    \item \textqcr{HRED-F}: Add the $\Delta_{fact}$ term to DBS \cite{DBS}, which generates the responses of HRED. $B'=1 (B=15, G=15)$.
    \item \textqcr{MHRED-F}: Add the $\Delta_{fact}$ term to DBS, which generates the responses of MHRED. $B'=1$ $(B=5, G=5)$.
    \item \textqcr{MHRED-F15-R, MHRED-F5-R}: Add the $\Delta_{fact}$ term to DBS, which generates the responses of MHRED. $B'=1$ $(B=15, G=15)$ or $(B=5, G=5)$. Reranker selects the final response from candidates returned by \textqcr{MHRED-F}.
    \item \textqcr{Ensemble}: Reranker selects the final response from candidates returned by both \textqcr{MHRED-F} and FR.
\end{itemize}

Moreover, the baseline models ``baseline(random)'' and ``baseline(constant)'' derived from the organizers are compared with the proposed model in human evaluation. 

Note that only FR model should not be compared with other models since FR model is not able to output responses continuously.

\subsection{Model Setup}
We use a two-layer seq2seq, HRED and MHRED for training. All models are set to the word embedding dimension and hidden vector size of 256. Mini-batch training was employed with a batch size of 40. The models were trained with cross entropy loss function and adapted Adam optimization algorithm \cite{adam} with the initial learning rate of 0.0001. To alleviate over-fitting to the training dataset, a dropout rate of 0.2 was set for all models. Training was conducted for up to 20 epochs and the model with the lowest perplexity in the dev dataset was selected.

Hyper-parameters of DBS was set as $\lambda = 0.4$ and $\gamma = 10.0$ according to BLEU on dev dataset.
Vocabulary size was set to 20k, which is shared between both the dialogue and facts data. In generating responses, the log probability of out-of-vocabulary (OOV) words was set to $-\infty$ so as not to generate the special symbol $<$unk$>$.

\begin{table}[t]
    \centering
    \footnotesize
    \begin{tabular}{c|cccc} \hline 
        \textbf{Model} & \textbf{NIST4} & \textbf{BLEU4} & \textbf{METEOR} & \textbf{div1}\\
        \hline \hline
        \textqcr{S2S} & 0.023 & 0.34& 3.92 & 0.026 \\
        \textqcr{HRED} & 0.730 & 0.58 & 5.65 & 0.049 \\
        \textqcr{HRED-F} & 0.766 & 0.68 & 5.61 & 0.049 \\
        \textqcr{MHRED-F} & 0.555 & 0.76 & 5.24 & 0.069 \\
        \textqcr{MHRED-F15-R} & 1.802 & 0.92 & 6.45 & 0.058  \\
        \textqcr{MHRED-F5-R} & 1.749 & 1.10 & \textbf{6.74} & 0.051  \\
        \textqcr{Ensemble} & \textbf{2.047} & \textbf{1.35} & 6.71 & \textbf{0.094} \\     \hline
    \end{tabular}
\caption{Results of the automatic evaluation.}
    \label{automatic}
\end{table}

\begin{table}[t]
    \centering
    \footnotesize
    \begin{tabular}{c|cc} \hline
        \textbf{Model} & \textbf{Appropriateness} & \textbf{informativeness} \\
        \hline \hline
        baseline(constant) & 2.60 & 2.32 \\ 
        baseline(random) & 2.32 & 2.35  \\ \hline
        \textqcr{Ensemble} & \textbf{2.69}  & \textbf{2.58} \\ \hline
    \end{tabular}
\caption{Results of the human evaluation.}
    \label{human}
\end{table}

\section{Results and Discussion}

\begin{table}[t]
    \centering
    \begin{tabular}{c|c|cc} \hline 
        \multirow{2}{*}{\textbf{Category}} & \multirow{2}{*}{\textbf{Feature}} &   \multicolumn{2}{c}{\textbf{Difference}} \\
        & & \textbf{Category} &  \textbf{Feature} \\ \hline \hline
        \multirow{4}{*}{Candidate} & \multicolumn{1}{c|}{Length} & & $+0.0004$ \\
        & \multicolumn{1}{c|}{Fluency} & & $\bm{-0.1053}$ \\
        & \multicolumn{1}{c|}{POS} & & $-0.0038$ \\
        & \multicolumn{1}{c|}{Facts} & \multirow{-4}{*}{$\bm{-0.1039}$} & $-0.0021$ \\ \hline
        \multirow{8}{*}{Pair} & \multicolumn{1}{c|}{Word sim} & & $-0.0021$ \\
        & \multicolumn{1}{c|}{$N$-gram sim} & & $+0.0013$ \\
        & \multicolumn{1}{c|}{Length sim} & & $-0.0027$ \\
        & \multicolumn{1}{c|}{Embedding sim} & & $-0.0021$ \\
        & \multicolumn{1}{c|}{Sentimental sim} & & $-0.0023$ \\
        & \multicolumn{1}{c|}{POS sim} & & $-0.0004$ \\
        & \multicolumn{1}{c|}{Proper Noun sim} &  \multirow{-8}{*}{$-0.0155$} & $\pm 0.0000$ \\ 
        & \multicolumn{1}{c|}{Keyword sim} & & $\bm{-0.0203}$ \\\hline
        \multirow{1}{*}{Context} & \multicolumn{1}{c|}{Topic sim} & \multirow{1}{*}{$-0.0015$} & $-0.0015$ \\ \hline
    \end{tabular}
    \caption{Differences of accuracy computed by Reranker when the target feature is excluded. }
    \label{features_acc}
\end{table}

        \begin{table*}[t]
            \centering
            \footnotesize
            \begin{tabular}{c|p{6.15cm}|p{9.65cm}} \hline 
                \# & \textbf{Context} & \textbf{Response}  \\ \hline \hline
                \multirow{6}{0.1cm}{1} & \multirow{6}{6.15cm}{til there is enough water in lake superior to flood the entire landmasses of north and south america to a depth of 1 foot. it contains over 3 quadrillion gallons of fresh water. $\to$ the world's largest by volume and deepest lake is located in southern russia. lake baikal.}
                & \textqcr{S2S:} i'm not true.\\
                & & \textqcr{HRED:} we are there. i don't really count, but its not..\\
                & & \textqcr{HRED-F:} also, and i have been posted \textit{lake baikal} are on. the water lake .\\
                & & \textqcr{Ensemble (FR):} \textit{lake baikal} volume. so the answer is yes a bit more than the all the great lakes, and a little under twice the volume of \textit{lake superior}. \\
                & & \\\hline
        
                \multirow{6}{0.1cm}{2} & \multirow{6}{6.25cm}{til that president jimmy carter placed his peanut farm in a blind trust when he was elected president. after his term ended, he found that it had been badly mismanaged and he was \$1 million in debt. $\to$ god bless jimmy carter america's last honest president.} 
                & \textqcr{S2S:} what do you think, i think of. \\
                & & \textqcr{HRED:} not sure. the best really liked by far.\\
                & & \textqcr{HRED-F:} yeah. but that is a president \textit{obama}.\\
                & & \textqcr{Ensemble (MHRED):} \textit{obama} was also a bit of course he was president \textit{obama}. \textit{democrat}.\\
                & & \\ \hline
            \end{tabular}
            \caption{Example of a response. The token $\to$ indicates a change of turn.
            \textqcr{Ensemble ($\cdot$)} denotes the output source (either MHRED or FR) selected by the Reranker.}
            \label{response}
        \end{table*}
        \begin{table*}[t]
            \centering
            \footnotesize
            \begin{tabular}{c|p{5.35cm}|p{7.90cm}|l|l} \hline
                \# & \textbf{Context} & \textbf{Response} & \textbf{CS} & \textbf{Rank} \\ \hline \hline
                \multirow{4}{0.1cm}{1} & \multirow{4}{5.35cm}{til that kyoto, the former capital of japan, just means ``capital city'' and tokyo means ``eastern capital'' $\to$ i only just noticed that tokyo and kyoto are anagrams.}  &\textqcr{MHRED: }i think tokyo godzilla, but as well and kyoto. & 0.9602 & 1\\
               & & \textqcr{MHRED: }they also have been a lot of tokyo as the tokyo are they have the same as well. the kyoto is the only one. & 0.9342& 2\\
               & & \textqcr{FR: }villages arent cities.  & 0.1817 & worst\\ \hline 
                
                \multirow{5}{0.1cm}{2} & \multirow{5}{5.35cm}{til german animal protection law prohibits killing of vertebrates without proper reason. because of this ruling, all german animal shelters are no-kill shelters. $\to$ i am german. til that there are kill shelters.}  & \textqcr{FR: }wow! i didnt know there was a tv show about for pets/animal shelters. thats pretty cool! do you know if that sort of advertising caused a lot more people to adopt animals? & 0.8144 & 1\\
                &  & \textqcr{MHRED: }its a good thing about cats are occupying breeds cats. & 0.7297 & 2\\
                &  & \textqcr{FR: }use its hide as shelter. & 0.7234 & 3\\ \hline
            \end{tabular}
            \caption{Example of reranking by the Reranker. The token $\to$ indicates a change of turn.
            CS represents the confidence score produced by XGBoost.}
            \label{reranker_res}
            \end{table*}

Table \ref{tab:automatic} shows results of the automatic evaluation. It can be seen that the proposed model \textqcr{Ensemble} performs better than other models. This indicates that \textqcr{Ensemble} enables to output more fluent responses similar to human and diverse responses.

Comparing the result of \textqcr{MHRED-F} and \textqcr{HRED-F}, notably at div1 score, it is apparent that the proposed MHRED architecture is superior to conventional models. It shows that MHRED can infer words and topics using facts that may be hard to handle only from conversation data and generate diverse responses on the new domain.

Comparison of \textqcr{Ensemble} and \textqcr{MHRED-F5-R} indicates that the FR module is effective. This is because the responses by the FR are parts of the conversation actually chatted by human and thus highly fluent. Thus, it is shown that the MHRED and the FR are useful in generating informative and appropriate responses. 

To analyze effectiveness of introducing the penalty $\Delta_{fact}$ and the Reranker, we compared \textqcr{HRED-F}, \textqcr{HRED}, \textqcr{MHRED-F15-R} and \textqcr{MHRED-F}. The model combining the Reranker (\textqcr{MHRED-F15-R}) gives significantly higher performance than the model without the Reranker, even on a single model. It designates capturing diverse perspectives of dialogue with various features is important for response generation. Conversely, the model introducing the penalty $\Delta_{fact}$ (\textqcr{HRED-F}) showed slight improvement on NIST4 and BLEU4. This indicates that adding the penalty of DBS has positive potential to generate responses similar to human-made.

Table \ref{human} shows the results of human evaluation. Since our primary model beats official baseline models returning responses randomly and constantly, the proposed model is able to capture the context and generate responses fluently.

\section{Case Study and Error Analysis}

\begin{figure}[t]
    \centering
    \includegraphics[height=6.0cm, width=8.6cm]{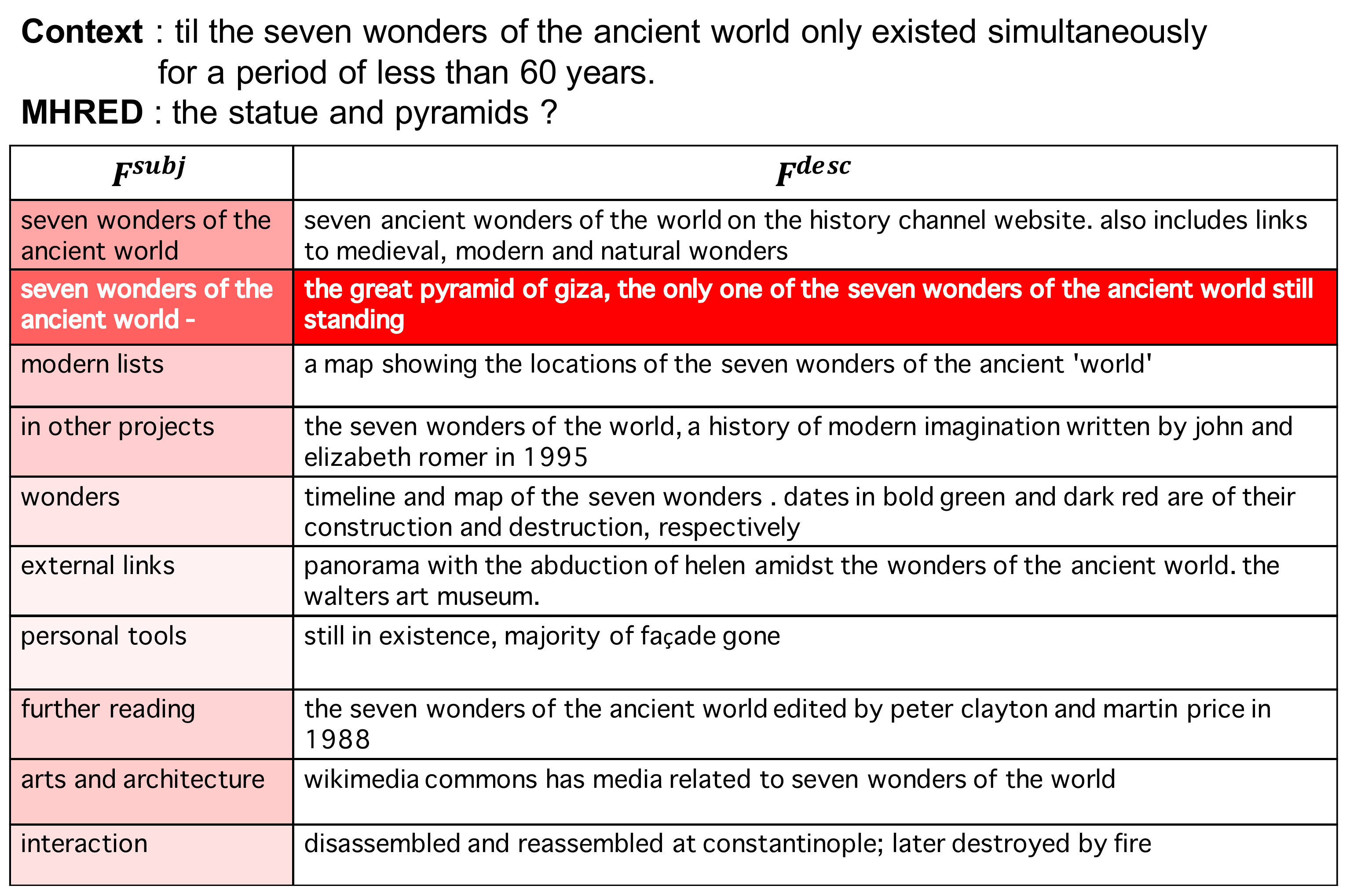}
    \caption{Attention paid by the facts encoder. Sentences painted in darker shades of red represent greater attention.}
    \label{attention}
\end{figure}

To validate the MHRED architecture, we looked into the details of the result with attention value $p^{subj}$ and $p^{desc}$ in the facts encoder. Figure \ref{attention} depicts an example of attention paid by the fact encoder. 
The $F^{subj}$ captures ``\textit{seven wonders of the ancient world}'', which refers to the topic of the context. Subsequently, The $F^{desc}$ captures the facts containing ``\textit{pyramid}'' considering both the context and $F^{subj}$. Finally, the MHRED generates a response including ``\textit{pyramid}". This indicates that this model enables to focus on the facts relevant to the context and generate responses injecting them.

Table \ref{features_acc} shows how the accuracy of the Reranker changes when one of the target features is excluded, per feature or per category.
A negative value implies that the corresponding feature is important. The category ``Candidate'' showed significant decrease of all categories, and the feature ``Fluency'' showed the biggest decrease by $-0.1053$, followed by ``Keyword sim'' by $-0.0203$. Thus, the Reranker has a tendency to select the final response focusing on fluency and contextually informativeness in dialogue. This tendency is probably due to making training dataset for the Reranker. Negative examples are generated using hand-crafted rules such as swapping and eliminating words, thus resulted in the tendency to select more higher ``Fluency'' and ``Keyword sim'' sentences preferentially.

Table \ref{response} shows examples of responses predicted by the models. As can be observed from the table, \textqcr{HRED-F} and \textqcr{Ensemble} output more informative words related to the context such as ``\textit{lake bikal}'' (\#1) or ``\textit{obama}'' (\#2). Table \ref{reranker_res} presents examples of reranking by the Reranker.
In example \#1, the MHRED is explicitly designed for the previous context, and the Reranker selects the most meaningful response. In example \#2, the response returned by the FR has high fluency and many content words. Conversely, the response is not suitable for the context in terms of the topic. This indicates, as above mentioned, that the Reranker tends to focus on ``Candidate'' strongly due to the way of making examples for the Reranker. However, we expects making examples from the various perspective will improve the performance more.

\section{Conclusion and Future Work}
In this paper, we proposed an ensemble dialogue system using external facts for DSTC7-Task2. The proposed system is a combination of three modules: the MHRED, a neural dialogue system which incorporates external facts into the procedure of response generation, the FR, and the Reranker. In generation, we extend the DBS to generate more meaningful words containing facts data. The experimental results showed that the MHRED especially improved the diversity of the response sentence over the baseline model. Moreover, we confirmed that the combination of multiple modules improved overall automatic metrics and generates more informative responses. In future work, we plan to introduce an end-to-end learning for multiple systems simultaneously.

\bibliographystyle{aaai.bst}
\footnotesize{

}
\end{document}